\documentclass[runningheads]{llncs}
\usepackage[table,xcdraw]{xcolor}
\usepackage{xcolor}
\usepackage{times}
\usepackage{graphicx}
\usepackage{amsmath}
\usepackage{amssymb}
\usepackage{svg}
\usepackage{amsmath}
\usepackage{booktabs}
\usepackage{bm}
\usepackage{upgreek}
\makeatletter
\DeclareRobustCommand\onedot{\futurelet\@let@token\@onedot}
\def\@onedot{\ifx\@let@token.\else.\null\fi\xspace}

\usepackage{graphicx}

\begin{document}
\title{Feature Selection on Thermal-stress Dataset}
\author{Xuyang Shen \and Jo Plested \and Tom Gedeon}
\authorrunning{X. Shen et al.}
\institute{Research School of Computer Science,\\ Australian National University \\
\email{first.second@anu.edu.au}}
\maketitle              % typeset the header of the contribution
\begin{abstract}
Physical symptoms caused by high stress commonly happen in our daily lives, leading to the importance of stress recognition systems. This study aims to improve stress classification by selecting appropriate features from Thermal-stress data, ANUstressDB. We explored three different feature selection techniques: correlation analysis, magnitude measure, and genetic algorithm. Support Vector Machine (SVM) and Artificial Neural Network (ANN) models were involved in measuring these three algorithms. Our result indicates that the genetic algorithm combined with ANNs can improve the prediction accuracy by 19.1$\%$ compared to the baseline. Moreover, the magnitude measure performed best among the three feature selection algorithms regarding the balance of computation time and performance. These findings are likely to improve the accuracy of current stress recognition systems.

\end{abstract}

\section{Introduction}

Stress as an emotional language of humans represents the body’s reaction to the environment. Light stress can be beneficial to our body~\cite{stress_suvery}, while high stress may let us feel less energetic, give us headaches, and even cause major illness. According to the latest report from the American Institute of Stress~\cite{stress_effect}, job stress affects 46$\%$ American adults and that it has escalated gradually over the past decades. Correctly judging employees’ stress not only helps leaders to adjust workload but also is favorable for the mental and physical health of employees. 

Traditional stress recognition and detection systems~\cite{chen2014detection} require physiological signals collected from special devices like blood pressure cuff, which is not suitable to be utilized in normal circumstances. Therefore, Irani et al.~\cite{irani2016thermal} proposed a real-time stress recognition system based on the physical appearance of objects, which requires an RGB camera and a thermal camera. In this paper, we focused on the same dataset as~\cite{irani2016thermal} and analyzed which feature selection algorithms can sufficiently improve the stress classification for Support Vector Machine(SVM) and Artificial Neural Network(ANN) respectively.

Reducing data sources from several physiological devices into two cameras improves the feasibility of the real application's recognition system but requires stricter data pre-processing. Feature selection as one pre-processing sub-task helps to filter out meaningless image information and hence improves the model performance~\cite{peng2005feature}. We first explore correlation analysis from information theory that aims to measure the strength of the linear relationship between features~\cite{james2013introduction}. We also employ the magnitude measure that is applied initially to prune neural networks~\cite{gedeon1997data}. It helps us to consider the feature relationship from learned neurons through local search. The genetic algorithm was introduced as the third method, which uses the stochastic search to approach global optimization~\cite{goldenberg1989genetic}.

\section{Method}

\subsection{Network Structure and Hyper-parameters}
\subsubsection{Support Vector Machine, SVM}

SVM is a supervised machine learning model that performs both linear and non-linear classification through the kernel trick~\cite{cortes1995support,boser1992training}. We explored it as a baseline to compare with ANN as it is successfully applied in various applications in computer vision~\cite{taini2008facial}. The SVM utilized in this paper was imported from Sklearn by using default hyper-parameters. To maintain the fairness of each generation in genetic algorithms, we set a fixed value in the random state.

\begin{table*}[]
    \centering
    \caption{Hyper-Parameters of SVM}
\begin{tabular}{ccccc}
    \hline
    \textbf{Kernel} & \textbf{Degree} & \textbf{Gamma} & \textbf{Degree} & \textbf{Random State} \\ \hline
    rbf             & 3               & auto           & 3               & 22                    \\ \hline
\end{tabular}
\end{table*}

\subsubsection{Artificial Neural Network, ANN}

\begin{figure}[t]
      \centering
      \includegraphics[height=4cm]{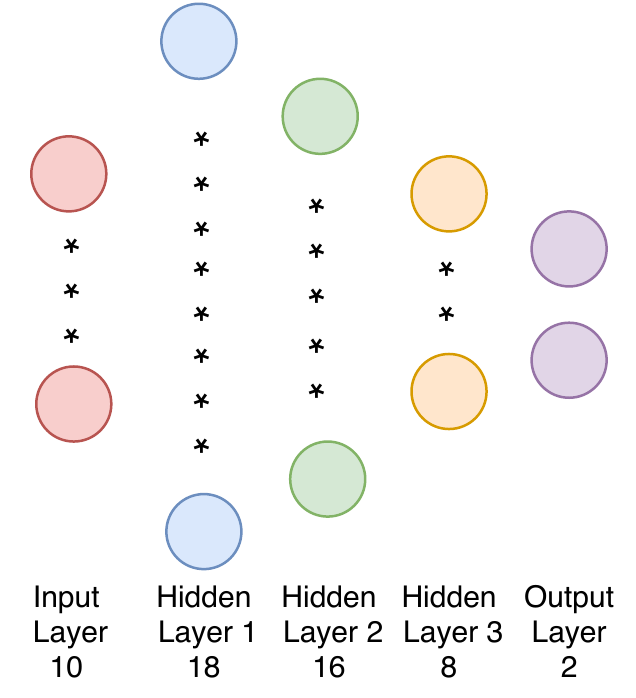}
      \caption{Four-Layer neural network (ANN).}
      \label{fig:ann_layer}
\end{figure}

We examined several architectures of fully connected neural networks (FCNN) to recognize the stress. The first network topology is 10-10-6-2, being 10 input neurons, 2 hidden layers, and 2 output neurons. However, it failed in generalizing to data, as we shall show in Section 3. Since the inputs are the features taken from raw pixels, a shallow neural network might struggle to capture the complex relationships between the pixels. The second one is 10-18-16-8-2 (Fig.~\ref{fig:ann_layer}), which adds a few neurons in each hidden layer and one extra hidden layer compared to the previous shallow FCNN.

One major issue of neural network design is how to optimize the network architecture to achieve a balance between overfitting and generalization~\cite{zhang2016understanding}. Considering that the shallow network cannot generalize well to thermal-stress data, we decided to use the second network topology in later experiments (Fig.~\ref{fig:ann_layer}). The hyper-parameter of ANN was also adjusted to improve the efficiency of training in genetic algorithms (Table~\ref{hyperparameters_ann}). For instance, Table~\ref{generlization_of_dataset} indicates both train-test separation and cross-validation can reflect similar test accuracy, but the former highly reduces the computation time. Adam is selected as the optimizer, which contributes to boosting the performance in the non-convex problem in a shorter time ~\cite{kingma2014adam}.

\begin{table*}[]
    \centering
    \caption{Hyper-Parameter of ANN}

    \label{hyperparameters_ann}
\begin{tabular}{cccc}
    \hline
    \textbf{Optimization}                                             & \textbf{Learning Rate}                                                                  & \textbf{Weight Decay}                                                                   & \textbf{Epochs}      \\ \hline
    Adam                                                              & 0.0008                                                                                  & 0.0003                                                                                  & 10000                \\ \hline
    \textbf{Validation}                                               & \textbf{\begin{tabular}[c]{@{}c@{}}Activation Function \\ of Hidden Layer\end{tabular}} & \textbf{\begin{tabular}[c]{@{}c@{}}Activation Function \\ of Output Layer\end{tabular}} & \textbf{Random Seed} \\ \hline
    \begin{tabular}[c]{@{}c@{}}Training: 0.7\\ Test: 0.3\end{tabular} & ReLU                                                                                    & SoftMax                                                                                 & 7                    \\ \hline
\end{tabular}
\end{table*}

\subsection{Feature Selection Techniques}
\subsubsection{Correlation Analysis}

Correlation indicates the dependence or relationship between two data variables from a statistical perspective. The result of correlation is located between -1 and 1, which represents the dependence between two variables from negative correlation into positive correlation. We applied the correlation analysis to perform feature selection by filtering highly correlated variables from the statistical property of the dataset.

\begin{equation}
    corr(X,Y) = \frac{cov(X,Y)}{\sigma_{x} \ \sigma_{y}} = \frac{E\left [(X - \mu_{X}) (Y - \mu_{Y})  \right ]}{\sigma_{x} \ \sigma_{y}}
\end{equation} 
where $E$ means the expected values, and $cov$ refers to convariance.

\subsubsection{Magnitude Measure}
Gedeon~\cite{gedeon1997data} proposed the magnitude measure of input neuron to output neuron based on the measure of input neuron to hidden layers ~\cite{wong1995improved}. It is a static analysis applied for the input neuron in trained ANN. Compared with the correlation analysis, the magnitude measure can analyze the weight between layers, which indicates the impact of input neurons to others. As a result, we can perform feature selection based on the ranking of input variables, giving us higher quality results in the experiments.

\begin{equation}
    Q_{ik} = \sum_{r=1}^{nh}(P_{ir} \times P_{rk}),\ 
\text{where}\ P_{ir}=\frac{\left | W_{ij} \right |}{{\sum_{p=1}^{ni}} \left |W_{pk} \right |} \
P_{jk}=\frac{\left | W_{jk} \right |}{{\sum_{r=1}^{nh}} \left |W_{rk} \right |}
\end{equation}
where $P_{ij}$ refers the influence of input neuron $i$ to the hidden neuron $j$, and $P_{jk}$ measures the effect of hidden neuron $j$ to the output neuron $k$.

\subsubsection{Genetic Algorithm}

The magnitude measure has large limitations which make it not applicable to SVM and highly non-deterministic in ANN as it performs the local search. Therefore, we employed the genetic algorithm to analyze the feature selection further.

\begin{figure}[t]
      \centering
      \includegraphics[height=6cm]{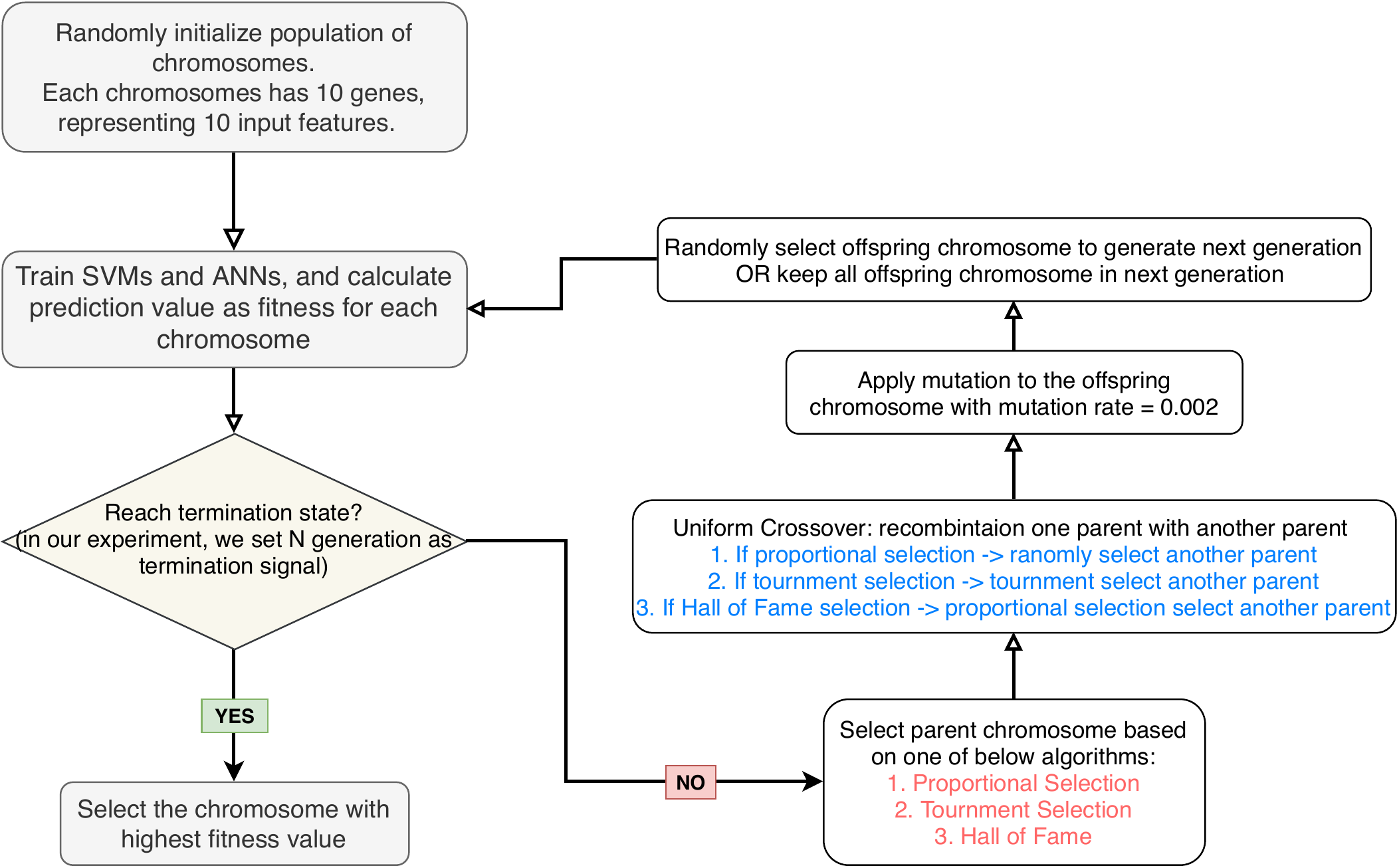}
      \caption{Combining SVM and ANN with Genetic Algorithm to explore feature selection}
      \label{fig:ga}
\end{figure}

Genetic algorithms, derived from natural selection~\cite{mitchell1998introduction}, are one large class of evolutionary algorithms, performing the stochastic search for an optimal solution based on 5 components (described in Fig.~\ref{fig:ga}). The fitness function is the main module among these 5 components, responsible for evaluating each chromosome (feature selection). Typically, hybrid values from train accuracy and validation accuracy of models are selected as fitness scores. In contrast, it is not suitable in our experiment since the model training on thermal-stress dataset can easily overfit (Table~\ref{generlization_of_dataset}). We only picked validation accuracy; meanwhile, we also set a fixed seed for all random states in model initialization and train-test dataset separation.

We also experimented with three different selection and crossover techniques, as described below:

\begin{itemize}
\item Use proportional selection to select one parent and randomly select the other parent. The principle of this selection algorithm is that chromosomes with large fitness values have a higher probability of passing their genes to offspring.
\begin{equation}
    \varphi_{s} (x_{i}(t)) = \frac{f_{r} (x_{i}(t)) }{\sum_{l=1}^{ns} (f_{r} (x_{l}(t)))}
\end{equation}
where $n_{s}$ is the total number of individuals in the population; $ \varphi_{s} (x_{i})$ is the probability that $x_{i}$ will be selected; $f_{r} (x_{i})$ is the scaled fitness value of $x_{i}$

\item Tournament selection for both parent: the best individual in the group of $n_{ts}$ chromosomes will become the parents (set $n_{ts}=0.6*\text{population}$). The principle of tournament selection is to limit the chance of the best individual to dominate.

\item Use Hall of Fame selection to select one parent and use the proportional selection to select the other parent; only the best individual of each generation is chosen to be inserted into the hall of fame, which becomes a parent pool for crossover operator.
\end{itemize}

\section{Dataset: Thermal-stress Dataset}

The ANUStressDB~\cite{irani2016thermal} dataset was generated from an HCI experiment about stress simulator, which involved 31 subjects. Their facial information was recorded by a Microsoft webcam and FLIR camera at 30 frames per second at 640x480 pixels~\cite{irani2016thermal}. The raw data is then PCA processed to protect privacy. 

The thermal-stress dataset contains 620 records with five features extracted from RGB facial images and five features extracted from the thermal camera as described above. Specifically, these ten features vectors are derived from the result of principal component analysis (PCA) on the original raw data. It also gives us a classification label of each record that is either stressful or calm. The labels were manually constructed in the experiment environment setting and validated by the questionnaire survey.

\begin{figure}[t]
      \centering
      \includegraphics[height=4cm]{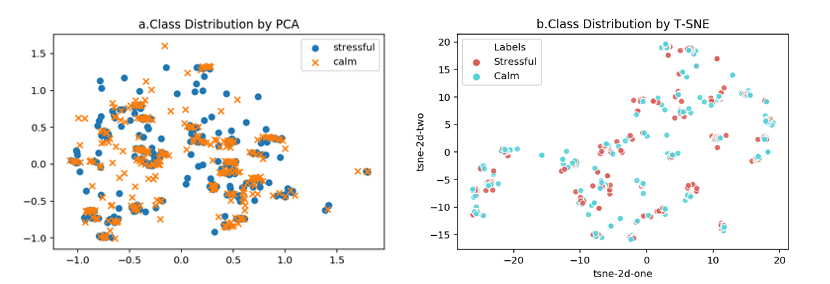}
      \caption{PCA and t-SNE dimensional reduction techniques are applied to visualize the original ten-dimension data}
      \label{fig:dataset_distribution}
\end{figure}

Although the dataset is well-balanced between two classes, it is still hard for the model to make a correct prediction. From two visualizations (Fig.~\ref{fig:dataset_distribution}), it indicates a high-class similarity between stressful and calm data. Most of the data points from two different classes are overlapped. Most spatial information is lost after the main feature extraction, which makes the prediction task more difficult.

The results of experiments also prove that the trained model fails in prediction (Table~\ref{generlization_of_dataset}). In the 10-18-16-8-2 model, high training accuracy indicates the networks learn several features from training data but still fail in the prediction. 

\begin{table*}[]
    \centering
    \caption{Training and Testing Accuracy of two models and validation methods (report the average result among 20 times running). 10-18-16-8-2: 10-input neurons; 18-hidden1 neurons; 16-hidden2 neurons; 8 hidden3 neurons, 2 output neurons. 10-10-6-2: 10-input neurons; 10-hidden1 neurons; 6-hidden2 neurons; 2 output neurons}
    \label{generlization_of_dataset}
\begin{tabular}{cccc}
    \hline
    \textbf{Model} & \textbf{Validation Method} & \textbf{Training Accuracy} & \textbf{Test Accuracy} \\ \hline
    10-18-16-8-2   & Training: 0.7 Test: 0.3    & 92.84\%                    & 53.50\%                \\
    10-18-16-8-2   & 5-Fold Cross-validation    & 91.71\%                    & 52.10\%                \\
    10-10-6-2      & Training: 0.7 Test: 0.3    & 67.3\%                     & 51.2\%                 \\
    10-10-6-2      & 5-Fold Cross-validation    & 63.2\%                     & 49.7\%                 \\ \hline
    \end{tabular}
\end{table*}

\section{Results and Discussion}
\subsection{Baseline of Feature Selection Experiment}
To reflect the influence of different feature selection algorithms, we computed the baseline of SVM and ANN model, which took all of ten features as inputs (Table~\ref{baseline}). The table's data was the result of one run since all random states were set in a fixed value.

\begin{table}[]
    \centering
    \caption{Baseline of SVM and ANN}
    \label{baseline}
\begin{tabular}{cccc}
    \hline
    \textbf{Model} & \textbf{Final Train Accuracy} & \textbf{Test Accuracy} & \textbf{Platform} \\ \hline
    SVM            & 100\%                         & 49\%                   & MacOS             \\
    ANN            & 95.33\%                       & 46.47\%                & Linux             \\ \hline
\end{tabular}
\end{table}

SVM with ten input features can reach 100$\%$ final training accuracy and 49$\%$ test accuracy. The evaluation accuracy of the ANN is 47$\%$ on Linux platforms.

\subsection{Correlation Analysis on Feature Selection}
\begin{figure}[t]
      \centering
      \includegraphics[height=2.5cm]{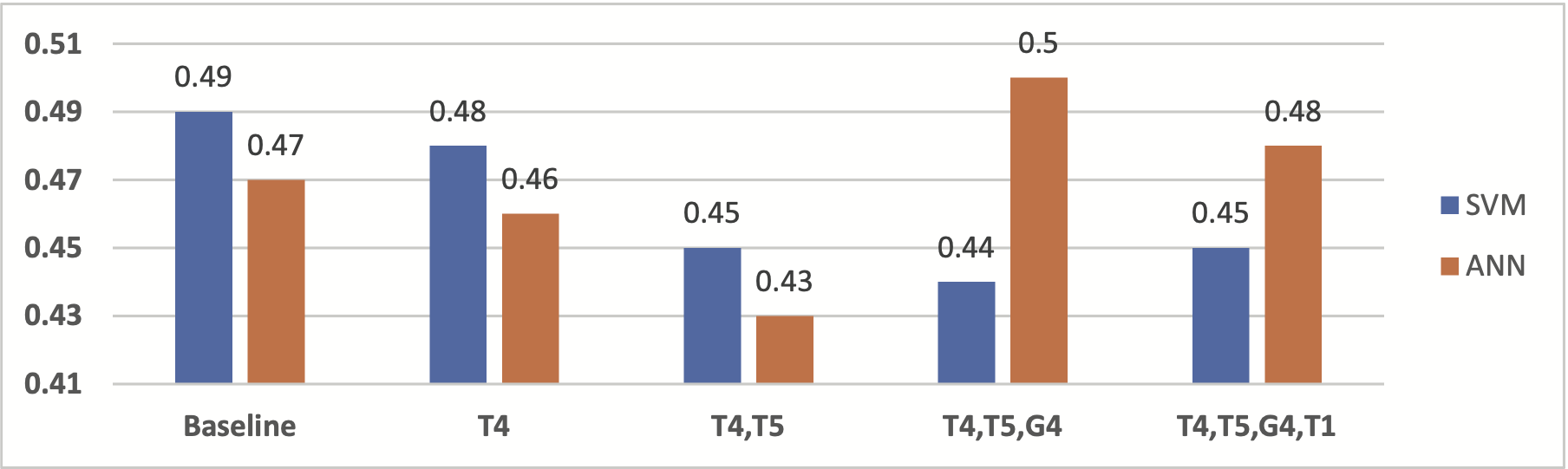}
      \caption{Test accuracy of removing the lowest correlated feature to top 4 lowest correlated features T4: remove Thermal\_4 feature; T4,T5: remove Thermal\_4 and Thermal\_5 features, and etc.  Besides, the training platform of ANN is Linux.}
      \label{fig:corrleation}
\end{figure}

The absolute values of overall correlation among these ten features are all lower than 0.1, and there are four attributes even lower than 0.005. To quantify the performance of correlation analysis, we further explored the influence of removing low correlated features to prediction accuracy (Fig.~\ref{fig:corrleation}). The result indicated that correlation analysis could not help gain better SVM performance, and it contributes little to the ANN as well. Therefore, we conclude that the correlation analysis cannot guide the feature selection in positive ways.

\begin{table*}[]
    \centering
    \caption{Correlation between different feature and labels, ranking from low correlation to high correlation}
    \label{correlation}
\begin{tabular}{cccccccccc}
    \hline
    \textbf{Thermal\_4} & \textbf{Thermal\_5} & \textbf{RGB\_4} & \textbf{Thermal\_1} & \textbf{RGB\_1} & \textbf{RGB\_2} & \textbf{RGB\_3} & \textbf{Thermal\_3} & \textbf{Thermal\_2} & \textbf{RGB\_5} \\ \hline
    -0.0015             & 0.0025              & 0.0039          & -0.0040             & 0.0054          & 0.0055          & -0.0093         & 0.010               & 0.015               & 0.023           \\ \hline
\end{tabular}
\end{table*}

\subsection{Magnitude Measure on Feature Selection}

\begin{figure}[t]
      \centering
      \includegraphics[height=2.5cm]{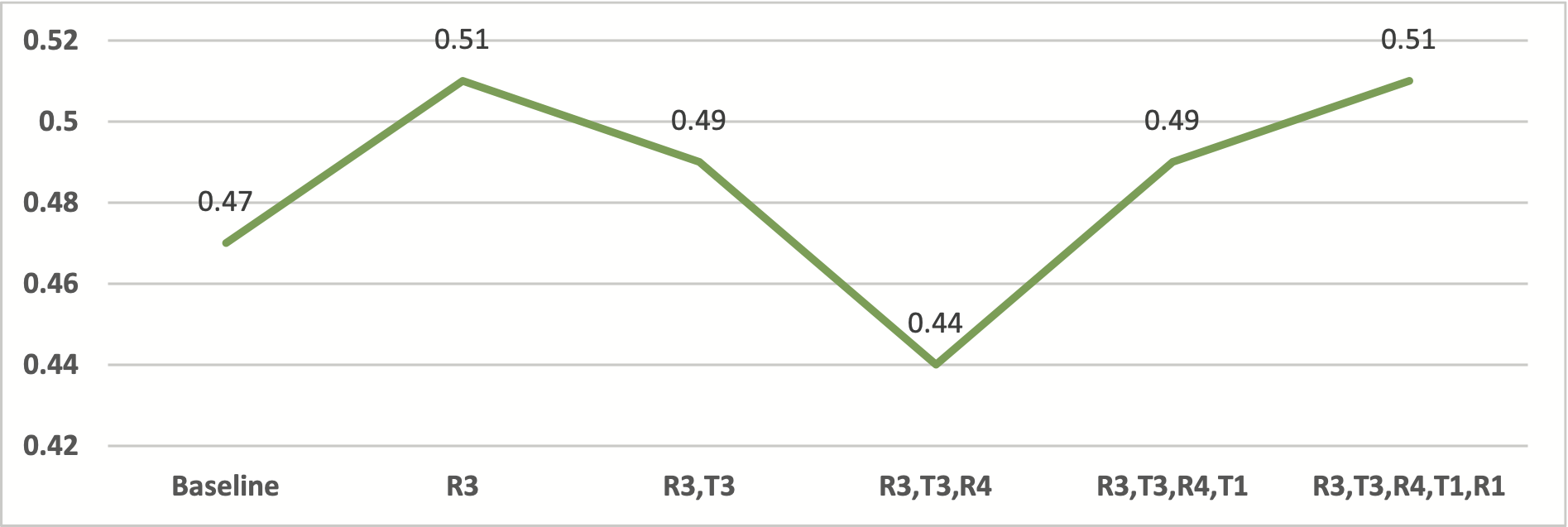}
      \caption{Test accuracy of removing least top1 – top 5 important variables. The training platform of ANN is Linux.}
      \label{fig:magnitude_measure}
\end{figure}

Compared to the correlation analysis, the magnitude measure focuses on the impact of dataset property on neural networks' behavior. We calculated the average magnitude value of input neurons behavior to the hidden neurons in 20 runs (Table~\ref{table:magnitude_measure}) and applied the results to feature selection experiments (Fig.~\ref{fig:magnitude_measure}).

The result indicated that "RGB\_3", "Thermal\_1", "RGB\_1" negatively contribute to the ANN's training and prediction, which were shown to be less important attributes in the magnitude measure. On the contrary, "RGB\_4" was incorrectly marked as less important data to hidden neurons by the magnitude measure. It was caused by the low stability of the magnitude measure. This method targets the weight's calculation of trained networks, which can be easily affected by the initial weights assigned to each hidden neurons and final termination judgment. Overall, the magnitude measure can provide valuable guidance to feature selection for ANN.

\begin{table*}[]
    \centering
    \caption{Magnitude of each attribute, ranking from low to high. To obtain a reasonable result, we set all random states in random values}
    \label{table:magnitude_measure}
\begin{tabular}{cccccccccc}
    \hline
    \textbf{RGB\_3} & \textbf{Thermal\_3} & \textbf{RGB\_4} & \textbf{Thermal\_1} & \textbf{RGB\_1} & \textbf{Thermal\_5} & \textbf{RGB\_2} & \textbf{Thermal\_4} & \textbf{Thermal\_2} & \textbf{RGB\_5} \\ \hline
    1.693           & 1.745               & 1.761           & 1.773               & 1.793           & 1.811               & 1.838           & 1.843               & 1.850               & 1.887 
    \\ \hline
\end{tabular}
\end{table*}

\subsection{Genetic Algorithm (GA) on Feature Selection}
We found the magnitude measure to be sensitive to the neuron weights through the previous experiments, which were updated through gradient descent in ANN's training. Backpropagation with gradient descent is a local search and is known to have local-minima issue. Besides, the magnitude measure is also limited to the learning model, which cannot be applied to SVM. Therefore, we also explored a global search to feature selection, which is genetic algorithms from evolutionary algorithms. Following the procedures introduced in section 2.2, we applied genetic algorithms to SVM and ANN, respectively, to further analyze the feature selection. Evaluating each chromosome with the validation accuracy of ANN requires 10 seconds, which means 60 population in one generation needs 10 minutes. Hence, we employed two computer resources, which causes a different baseline in Table ~\ref{baseline}. 

The result in Table ~\ref{ga1} indicated that GA applied with the ANN has a large improvement to the evaluation accuracy, which achieved 19.1$\%$ at maximum. It proved that stochastic global search with a high initial diversity has a higher probability of finding the best optimization. Additionally, the correlations between the attributes and the ground truths are low in the thermal-stress dataset, exaggerating the difference between local search and global search. The results of GA also showed that "RGB\_2", "RGB\_4", "Thermal\_2", and "Thermal\_3" were four important features.  

We also found that the population's initial diversity has different influences regarding different selection strategies of genetic algorithms. Specifically, the proportional selection was more sensitive to the initial diversity than then tournament selection. Hence, properly dominating superior genes is beneficial to the stability of genetic algorithms. 

Apart from applying genetic algorithms with ANN, we also explored the SVMs (Table~\ref{svm_ga}). Performing feature selection with genetic algorithms can improve 8$\%$ test accuracy of SVM at maximum. In conclusion, the genetic algorithm is the best option for either ANN models or SVM models. Not only it has broad applicability, but it also improves the model performance.

\begin{table}[]
    \centering
    \caption{GA algorithm applied with ANN. DNA order: [RGB\_1, RGB\_2, RGB\_3, RGB\_4, RGB\_5, Thermal\_1, Thermal\_2, Thermal\_3, Thermal\_4, Thermal\_5] 1 means this attribute is selected and vice versa}
    \label{ga1}
\begin{tabular}{c|ccccc}
    \hline
    \textbf{GA}                                                               & \multicolumn{2}{c}{\textbf{Proportional selection}}                                                    & \textbf{Hall of Fame selection}                  & \multicolumn{2}{c}{\textbf{Tournament selection}}                                                  \\ \hline
    \textbf{\begin{tabular}[c]{@{}c@{}}Population \\ Generation\end{tabular}} & \begin{tabular}[c]{@{}c@{}}80  \\  78\end{tabular} & \begin{tabular}[c]{@{}c@{}}60 \\ 100\end{tabular} & \begin{tabular}[c]{@{}c@{}}60\\ 100\end{tabular} & \begin{tabular}[c]{@{}c@{}}60\\ 100\end{tabular} & \begin{tabular}[c]{@{}c@{}}80\\ 78\end{tabular} \\ \hline
    \textbf{DNA}                                                              & 0101001101                                         & 0110111011                                        & 0001111110                                       & 1101011110                                       & 1111110011                                      \\
    \textbf{Test Accuracy}                                                    & 0.56                                               & 0.51                                              & 0.55                                             & 0.54                                             & 0.53                                            \\
    \textbf{Baseline}                                                         & 0.47                                               & 0.47                                              & 0.47                                             & 0.47                                             & 0.47                                            \\
    \textbf{Improvement}                                                      & \cellcolor[HTML]{C5E0B3}+0.09                      & \cellcolor[HTML]{C5E0B3}+ 0.04                    & \cellcolor[HTML]{C5E0B3}+0.08                    & \cellcolor[HTML]{C5E0B3}+ 0.07                   & \cellcolor[HTML]{C5E0B3}+ 0.06                  \\ \hline
\end{tabular}
\end{table}

\begin{table*}[]
    \centering
    \caption{GA algorithm applied with SVM. DNA order: [RGB\_1, RGB\_2, RGB\_3, RGB\_4, RGB\_5, Thermal\_1, Thermal\_2, Thermal\_3, Thermal\_4, Thermal\_5] 1 means this attribute is selected and vice versa}
    \label{svm_ga}
\begin{tabular}{c|ccc}
    \hline
    \textbf{GA}                                                              & \textbf{Proportional   selection}                & \textbf{Hall   of Fame selection}                & \textbf{Tournament   selection}                  \\ \hline
    \textbf{\begin{tabular}[c]{@{}c@{}}Population\\ Generation\end{tabular}} & \begin{tabular}[c]{@{}c@{}}60\\ 100\end{tabular} & \begin{tabular}[c]{@{}c@{}}60\\ 100\end{tabular} & \begin{tabular}[c]{@{}c@{}}60\\ 100\end{tabular} \\ \hline
    \textbf{DNA}                                                             & 0111010010                                       & 0111111011                                       & 0101100011                                       \\
    \textbf{Test Accuracy}                                                   & 0.52                                             & 0.53                                             & 0.50                                             \\
    \textbf{Baseline}                                                        & 0.49                                             & 0.49                                             & 0.49                                             \\
    \textbf{Improvement}                                                     & \cellcolor[HTML]{C5E0B3}+0.03                    & \cellcolor[HTML]{C5E0B3}+0.04                    & \cellcolor[HTML]{C5E0B3}+ 0.01                   \\ \hline
\end{tabular}
\end{table*}

\section{Conclusion and Future work}
Feature selection as one of the essential parts of data pre-processing can improve the prediction accuracy for both SVN and ANN. Different feature selection algorithms may contribute differently to the improvement of model performance. As we explored in the experiments, correlation analysis focuses on the dataset's statistical property to provide selection guidance, which takes the least computation power but is sensitive to the dataset. Since the correlation between attributes and ground truths is low in the thermal-stress data, this algorithm cannot guide sufficient information for us to choose features. The second algorithm we explored is the magnitude measure, which is moderate among the three algorithms we discussed in section 3, with a balanced ratio of performance and efficiency. However, the same as correlation analysis, magnitude measure only provides feature selection guidelines to us, and we have to decide which feature to keep or remove manually. Eventually, genetic algorithms can select the most appropriate features for both SVM and ANN but require huge computation power. For instance, 1,000 generations with 100 initial population of genetic algorithm with ANN need ten days to simulate the evolution. We also discovered that the genetic algorithm with tournament selection is less sensitive to population and generation, helping us to gain an ideal performance in a limited time.
We leave the exploration to find the right balance between time and performance in feature selection by the genetic algorithm as future work. Apart from different selection or crossover configuration of genetic algorithm, we can also consider this from the perspective of fitness functions.

\subsection*{Acknowledgements}
We thank Alasdair Tran for his invaluable feedback on the draft of this paper.


\end{document}